\documentclass[runningheads]{llncs}

\usepackage[mobile]{eccv}

\usepackage{eccvabbrv}
\usepackage{amsmath}
\usepackage{amssymb}
\usepackage{multirow}
\usepackage{colortbl}
\usepackage{mathtools}

\usepackage{graphicx}
\usepackage{booktabs}
\usepackage{colortbl}

\usepackage[accsupp]{axessibility}  

\usepackage{orcidlink}
\usepackage{multirow}
\DeclareMathAlphabet\mathbfcal{OMS}{cmsy}{b}{n}

\begin{document}

\title{TopoMaskV2: Enhanced Instance-Mask-Based Formulation for the Road Topology Problem
\vspace{-0.2cm}
} 

\titlerunning{TopoMaskV2}


\author{Muhammet Esat Kalfaoglu\orcidlink{0000-0001-5942-0454} \and
Halil Ibrahim Ozturk \and
Ozsel Kilinc\orcidlink{0000-0002-5269-2382} \and Alptekin Temizel\orcidlink{0000-0001-6082-2573}}

\authorrunning{E. Kalfaoglu et al.}

\institute{Graduate School of Informatics, METU, Ankara, Turkey 
\email{\{esat.kalfaoglu,atemizel\}@metu.edu.tr}
}

\maketitle

\begin{abstract}
\vspace{-0.6cm}

Recently, the centerline has become a popular representation of lanes due to its advantages in solving the road topology problem. To enhance centerline prediction, we have developed a new approach called TopoMask. Unlike previous methods that rely on keypoints or parametric methods, TopoMask utilizes an instance-mask-based formulation coupled with a masked-attention-based transformer architecture. We introduce a quad-direction label representation to enrich the mask instances with flow information and design a corresponding post-processing technique for mask-to-centerline conversion. Additionally, we demonstrate that the instance-mask formulation provides complementary information to parametric Bezier regressions, and fusing both outputs leads to improved detection and topology performance. Moreover, we analyze the shortcomings of the pillar assumption in the Lift Splat technique and adapt a multi-height bin configuration. Experimental results show that TopoMask achieves state-of-the-art performance in the OpenLane-V2 dataset, increasing from 44.1 to 49.4 for Subset-A and 44.7 to 51.8 for Subset-B in the V1.1 OLS baseline. 
\vspace{-0.2cm}
\vspace{-0.5cm}
\end{abstract}

\section{Introduction}
\vspace{-0.2cm}
\label{sec:intro}

For a fully autonomous driving system, it is imperative not only to detect stationary entities such as lanes, traffic signs, and traffic lights but also to comprehend their interrelationships. The challenge of detecting and understanding the relationships between stationary objects is referred to as the \textit{road topology} problem. For example, without an understanding of road topology, the interactions of lane merges and splits cannot be accurately defined, resulting in a deficient comprehension of autonomous driving scenes. Furthermore, certain traffic lights or signs apply exclusively to specific lanes rather than to all lanes on the road, underscoring the importance of road topology understanding.

Centerlines provide a natural representation of lanes and are a critical component in addressing the road topology problem. While lane divider representation is predominant in the lane detection literature \cite{pan2018spatial, tabelini2021keep, liu2021condlanenet, tabelini2021polylanenet, feng2022rethinking, chen2022persformer, luo2023latr, wang2023bev, liao2022maptr, chen2023efficient}, it can be argued that centerlines are more suitable for association problems compared to lane dividers, thereby facilitating efficient post-processing techniques for road topology. Additionally, centerlines offer a more intuitive representation of the direction of traffic flow.

In this study, we aim to enhance the performance of centerline detection within the broader context of road topology understanding. We propose a novel method, termed \textit{TopoMask}, which employs an instance-mask formulation in the BEV (Bird’s Eye View) domain to represent centerlines. This approach diverges from the traditional keypoint-based methods \cite{liao2022maptr, liu2022vectormapnet, li2023lanesegnet, wang2023bev} and parametric methods \cite{li2023graphbased, wu2023topomlp, bai2023curveformer, qiao2023end} commonly used for polyline structures in the BEV domain. Although instance-mask formulation has been implicitly utilized for polylines in the BEV domain for query generation \cite{hu2023planning} or as an auxiliary loss \cite{li2023lanesegnet}, a detailed analysis of directly generating polylines from instance masks and comparing these methodologies is absent in the literature. Concurrently with the TopoMask challenge report \cite{kalfaoglu2023topomaskinstancemaskbasedformulationroad}, the MachMap challenge report \cite{qiao2023machmap} indicates the generation of polylines from instance masks but lacks detailed polyline generation processes and numerical performance comparisons with other methodologies. Furthermore, centerlines uniquely represent flow information, unlike other polyline structures. Consequently, standard instance-mask formulations in the BEV domain cannot be directly adapted for centerline prediction. The TopoMask architecture addresses this limitation by enriching instance-mask formulations with flow information (flow-aware instance masks) through a proposed \textit{quad-direction label representation}. To the best of our knowledge, TopoMask is the first study to employ instance-mask formulation for centerline prediction and integrate it into the topology pipeline.

Moreover, the TopoMask architecture incorporates both an instance mask head and a Bezier regression head derived from the same instance query, facilitating the integration of these outputs during the post-processing phase. To the best of our knowledge, this study is pioneering in its fusion of disparate output types for polyline prediction. The proposed fusion scheme, which combines mask and Bezier information, significantly enhances both detection accuracy and relational performance. Additionally, we re-evaluate the pillar assumption inherent in the Lift Splat technique \cite{Philion2020}, introducing a novel multi-height bin modification. This modification substantially improves centerline prediction and augments the understanding of road topology.

Furthermore, TopoMask employs a masked-attention transformer architecture within the BEV domain, distinguishing itself from the prevalent use of deformable attention for polyline structures \cite{cheng2022masked, li2023graphbased, wu2023topomlp, chen2022persformer, liao2022maptr, luo2023latr, li2023lanesegnet, liu2022vectormapnet, bai2023curveformer, zhou2024himap, choi2024mask2map, liu2024leveraging, xu2024insmapper, yu2023scalablemap}. This approach offers significant advantages in terms of computational efficiency and convergence compared to standard cross-attention mechanisms. Despite the relatively rare and recent application of masked attention \cite{qiao2023end, ding2023pivotnet, choi2024mask2map, zhou2024himap} in the BEV domain for polyline representations, our work advances this field by conducting a comprehensive numerical analysis of these attention mechanisms in the decoder structure, specifically focusing on centerline representations.

Experimental evaluations show that TopoMask has state-of-the-art performances in the OpenLane-V2 dataset, such that the performance increases from 44.1 to 49.4 for Subset-A, from 44.7 to 51.8 for Subset-B in the V1.1 OLS; from 28.5 to 34.5 for Subset-A, from 25.2 to 41.6 in Subset-B in the V1.1 DET$_{l}$ metrics. Our study has been recognized with an innovation award in the OpenLane Topology Challenge 2023 at CVPR \cite{kalfaoglu2023topomaskinstancemaskbasedformulationroad}.
The main contributions of this paper are as follows: 
\begin{itemize}
    \item To the best of our knowledge, this study is the first to provide a detailed analysis of directly generating polylines from instance masks in the BEV domain and to compare this approach with other polyline representations, such as the Bezier representation.
    \item The proposed quad-direction label representation provides instance masks with flow information (flow-aware instance masks) and enables centerline prediction from the instance mask formulation. 
    \item The proposed fusion of bezier regression and mask prediction head outputs increases the performance for both centerline prediction and topology relations. 
    \item Enhancing LSS \cite{Philion2020} with a multi-height bin implementation improves detection and topology performance, addressing the shortcomings of the pillar assumption in LSS.
    \item Although masked attention mechanisms have been employed for BEV HDMap element prediction, including polyline representations, there remains a notable absence of comprehensive numerical analyses comparing these mechanisms with other attention types. This study aims to fill this gap by evaluating the performance of masked attention in comparison to deformable attention and self-attention, with a particular emphasis on centerline prediction. To the best of our knowledge, this study is the first to employ masked attention for centerline prediction.
    \item Potential suggestions for improving the OpenLane-V2 V1.1 metric, aiming to enhance its practical applicability, are discussed. 
\end{itemize}

\vspace{-0.6cm}
\section{Related Work}
\vspace{-0.2cm}

\subsection{Segmentation Methods on the BEV domain}
\vspace{-0.2cm}
Semantic segmentation within the Bird’s Eye View (BEV) domain has been extensively explored in numerous studies \cite{Philion2020, Li2022b, xie2022m2bevmulticamerajoint3d, zhou2022cross, Peng2022}, addressing both dynamic and static objects, such as vehicles, lane dividers, and drivable areas. A significant limitation of semantic segmentation is its inability to distinguish between different instances. Fiery \cite{Hu2021}, inspired by Panoptic Deeplab \cite{cheng2020panopticdeeplabsimplestrongfast}, detects dynamic objects using an instance segmentation approach within the BEV domain. MapTRV2 \cite{liao2023maptrv2}, and GeMap \cite{zhang2023online} employ semantic segmentation as an auxiliary loss. FipTR \cite{gui2024fiptr} utilizes a transformer-based instance mask formulation to match vehicles across different time frames. Drawing from MaskDINO \cite{Li2022maskdino}, MachMap \cite{qiao2023machmap} adapts the instance-mask formulation to MapTR \cite{liao2022maptr} for detecting polyline structures in the BEV domain, but it lacks a comprehensive comparative analysis of instance mask formulation with other methodologies. LaneSegNet \cite{li2023lanesegnet} employs transformer-based BEV instance segmentation as an auxiliary loss, utilizing a mask head similar to TopoMask. UniAD \cite{hu2023planning} also adopts a transformer-based instance mask formulation to generate queries for road map elements, inspired by Panoptic Segformer \cite{li2022panopticsegformerdelvingdeeper}.  Diverging from existing literature, our study, TopoMask, aims to directly predict polylines from instance mask representation through a designed post-processing method rather than using an instance mask head for auxiliary loss or query generation. This approach facilitates the fusion with other polyline representations, such as the Bezier representation.

\vspace{-0.3cm}
\subsection{Masked Attention Concept on the BEV Domain for Polyline Prediction}
\vspace{-0.2cm}
The concept of masked attention, as proposed by the Mask2former \cite{cheng2022masked} in the PV domain, has demonstrated faster convergence and enhanced performance. In the TopoMask challenge report \cite{kalfaoglu2023topomaskinstancemaskbasedformulationroad}, we adapted this concept to the BEV domain for centerline prediction. Concurrently, BeMapNet \cite{qiao2023end} employs this concept within their transformer decoder for predicting Bezier coefficients of map elements. PivotNet \cite{ding2023pivotnet} integrates the masked attention mechanism with the point query concept, merging point queries into instance queries to implement masked attention. HIMap \cite{zhou2024himap} introduces a hybrid query representation that combines point and instance query paradigms. It applies deformable attention to point queries and masked attention to instance queries, facilitating interaction between these queries through a hybrid block. Mask2Map \cite{choi2024mask2map} leverages masked attention to enhance query features, utilizing these queries in a deformable-attention-based transformer decoder. Distinct from existing literature and in addition to our challenge report, we have conducted a comparative analysis of different attention types within the same architecture and methodology, with a particular focus on centerline prediction. To the best of our knowledge, this study is the first to analyze the performance of masked attention for centerline prediction and the road topology problem.
  
\vspace{-0.3cm}
\subsection{HDMap Element Prediction}
\vspace{-0.2cm}

The task of predicting High-Definition Map (HDMap) elements encompasses the detection of lane dividers, road dividers, and pedestrian crossings. HDMapNet \cite{li2022hdmapnet} transforms Bird’s Eye View (BEV) semantic segmentation into BEV instance segmentation through extensive post-processing, leveraging predicted instance embeddings and directional information. In contrast, VectorMapNet \cite{liu2022vectormapnet} employs a two-stage, coarse-to-fine methodology in an autoregressive manner, utilizing a map element concept. Diverging from the autoregressive approach, MapTR \cite{liao2022maptr} directly predicts points on polylines or polygons using a permutation-invariant Hungarian matcher, integrating point query and instance query concepts. InstaGraM \cite{shin2023instagram} redefines the polyline detection problem as a graph problem, where keypoints on the polyline are treated as vertices and the connections between keypoints as edges. MGMap \cite{liu2024mgmap} employs instance masks to generate map element queries and further refine features using mask outputs. MapVR \cite{zhang2024online} rasterizes the outputs of MapTR and applies instance segmentation loss to mitigate the limitations of keypoint-based methods. ADMap \cite{hu2024admap} utilizes instance interactive attention and vector direction difference koss to reduce point sequence jitter and improve map accuracy and stability. BeMapNet \cite{qiao2023end} models map elements as multiple piecewise curves, each represented using Bézier curves. StreamMapNet \cite{yuan2024streammapnet} investigates the temporal aspects of HDMap element prediction, utilizing temporal information as propagated instance queries and warped BEV features in a recurrent manner. MapTracker \cite{chen2024maptracker} enhances this concept by maintaining multiple memory latent rather than a single one and integrating them in a stridden fashion. 


\vspace{-0.3cm}
\subsection{Road Topology Problem and Centerline Concept}
\vspace{-0.2cm}

Road topology encompasses the interrelationships among lanes, as well as their connections with traffic lights and signs. However, the use of lane dividers proves to be an inefficient representation for addressing the topology problem, as each lane necessitates the use of two separate lane dividers. Consequently, the concept of centerlines has emerged as a more efficient and natural representation of lanes.

\vspace{-0.3cm}
\subsubsection{Centerline Concept}
\vspace{-0.2cm}
STSU \cite{Can2021} is one of the pioneering studies that predict centerlines instead of lane dividers. CenterLineDet \cite{xu2022centerlinedet} focuses on predicting centerlines from multiple cameras by leveraging temporal information and camera-LiDAR sensor fusion. LaneGAP \cite{liao2023lane} considers centerlines path-wise rather than piece-wise by preprocessing the centerline graph. MapTRV2 \cite{liao2023maptrv2} adapts MapTR for centerline prediction and improves efficiency. SMERF \cite{luo2023augmenting} aims to fuse the information of HDMap and multi-camera inputs. LaneSegNet \cite{li2023lanesegnet} is the first study extending the architecture towards the lane segment concept, which is an extension of the centerline concept that aims to predict also the boundaries of the lane.
\vspace{-0.3cm}
\subsubsection{Road Topology}
\vspace{-0.2cm}
TopoNet \cite{li2023graphbased} represents a pioneering study that proposes an architecture specifically designed to address the road topology problem. This study introduces a novel relational model that constructs a graph neural network, which interconnects centerlines with each other and with traffic elements while incorporating prior relational knowledge from preceding layers. Complementarily, TopoMLP \cite{wu2023topomlp} establishes a robust baseline for the road topology problem and proposes enhanced practices for addressing the relational aspects of this issue. CGNet \cite{han2024continuity} improves the continuity of centerline graphs and topology accuracy by incorporating three pivotal modules: Junction Aware Query Enhancement, Bézier Space Connection, and Iterative Topology Refinement.

\vspace{-0.4cm}
\section{Methodology}
\vspace{-0.2cm}

\begin{figure}[tb]
  \centering
  \includegraphics[width=\linewidth]{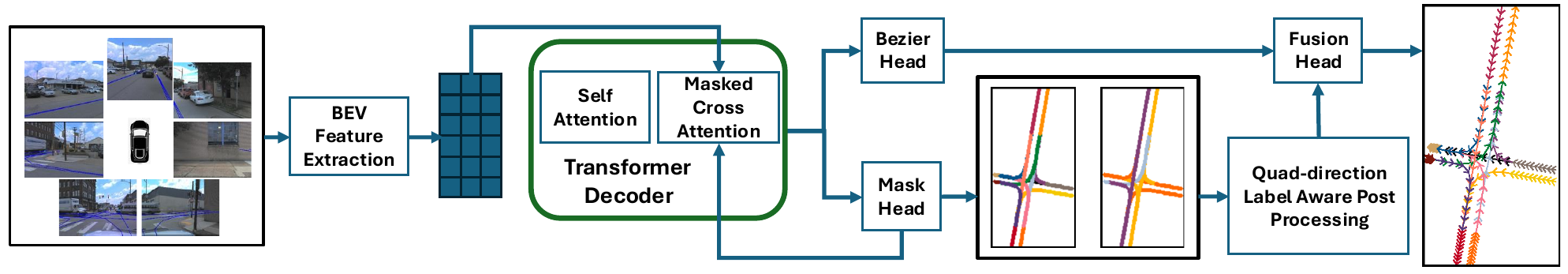}
  \vspace{-0.5cm}
  \caption{\textbf{Overview of the Centerline Prediction in the TopoMask Architecture.} The process starts with Bird’s Eye View (BEV) Feature Extraction, projecting features from multi-view images onto a unified coordinate system. Then, masked-attention-based transformer decoder creates flow-aware instance masks. These masks are converted into centerlines through a quad-direction label-aware post-processing step. To further improve centerline prediction, the Bezier Head is incorporated into the transformer decoder, and a fusion block is used to combine both representations.}
  \label{fig: topomask_architecture_centerline}
  \vspace{-0.5cm}
\end{figure}

The \textbf{TopoMask architecture} is a comprehensive road topology framework comprising four key branches: centerline detection, traffic element detection, centerline-centerline relationship estimation, and centerline-traffic element relationship estimation. The primary innovation of TopoMask lies in its centerline detection mechanism. This architecture employs a mask-based approach to represent centerline instances, capitalizing on the robustness and flexibility of segmentation techniques. Specifically, the centerline detection component utilizes an instance-mask formulation, where each centerline instance is depicted as a \textbf{flow-aware mask}. Each flow-aware mask includes a BEV (Bird’s Eye View) instance mask and a novel \textbf{quad-direction label representation} that encodes centerline flow information.

The centerline prediction component of the TopoMask architecture is illustrated in Figure \ref{fig: topomask_architecture_centerline}. The pipeline initiates with Bird’s Eye View (BEV) feature extraction, employing the Lift Splat (LSS) algorithm \cite{Philion2020}, which is further refined by transforming the pillar assumption into multi-height bins, akin to the BevFormer Architecture \cite{Li2022b}. These BEV features are subsequently processed as 2D image features. In contrast to existing topology baselines such as TopoMLP \cite{wu2023topomlp} and TopoNet \cite{li2023graphbased}, which utilize naive deformable attention, TopoMask employs masked attention \cite{cheng2022masked} to generate mask-aware queries for centerline prediction. Furthermore, TopoMask is augmented with a Bezier regression head. The predicted instance masks indirectly facilitate the prediction of Bezier control points through the masked attention mechanism, which concentrates attention on the regions of the predicted mask instances. Following the mask head, a quad-direction label-aware post-processing block is integrated. This block transforms each flow-aware instance mask into a 3D ordered point set, treating each centerline instance as an ordered 3D point set. Ultimately, a fusion block is incorporated to amalgamate the strengths of both the mask head and the Bezier head for centerline predictions.

\vspace{-0.4cm}
\subsection{Transformer Decoder in TopoMask Architecture}
\vspace{-0.2cm}

\begin{figure}[tb]
  \centering
  \includegraphics[width=0.99\linewidth]{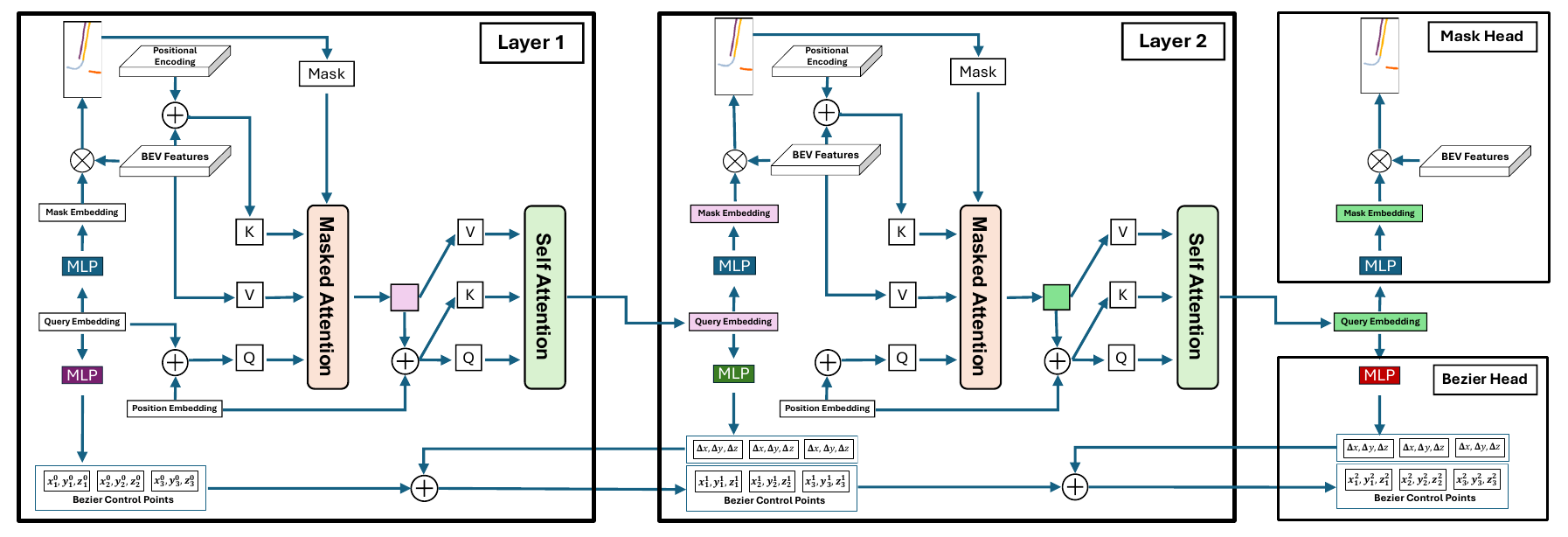}
  \vspace{-0.3cm}
  \caption{\textbf{Transformer Decoder in TopoMask} The TopoMask architecture comprises a mask head and a Bezier head, updating queries through successive layers of masked attention and self-attention within each decoder layer. Bezier control points are iteratively refined in each decoder layer, with the final binary mask output generated via a dot product between mask embeddings and BEV features.}
  \label{fig: topomask_masked_attention_detail}
  \vspace{-0.2cm}
\end{figure}

The transformer decoder in the TopoMask architecture leverages the masked-attention mechanism \cite{cheng2022masked}. This mechanism restricts attention to features within the predicted masked regions rather than learning attention weights for all features. Figure \ref{fig: topomask_masked_attention_detail} illustrates the transformer decoder of the TopoMask architecture. The inputs to the transformer decoder include BEV features, learnable query embeddings, and learnable positional embeddings. In the initial layer, preliminary predictions of masks and Bezier control points are generated. A Multi-Layer Perceptron (MLP) is employed to create mask embeddings from query embeddings, Final mask outputs are generated from the dot product of each mask embedding and the BEV features, followed by a sigmoid function. These BEV instance masks focus the attention weights on the predicted mask areas. Another MLP is used to obtain Bezier control points in a normalized manner. Each layer features self-attention followed by masked cross-attention. In subsequent layers, Bezier control points are iteratively updated, similar to the box regression approaches in DAB-DETR \cite{Liu2022_dab} and Deformable-DETR \cite{Zhu2020}.

\vspace{-0.3cm}
\subsection{Quad-direction Label Representation}
\vspace{-0.2cm}
\label{sec: quad_direction_label_representation}

\begin{figure}[tb]
  \centering
  \includegraphics[width=0.7\linewidth]{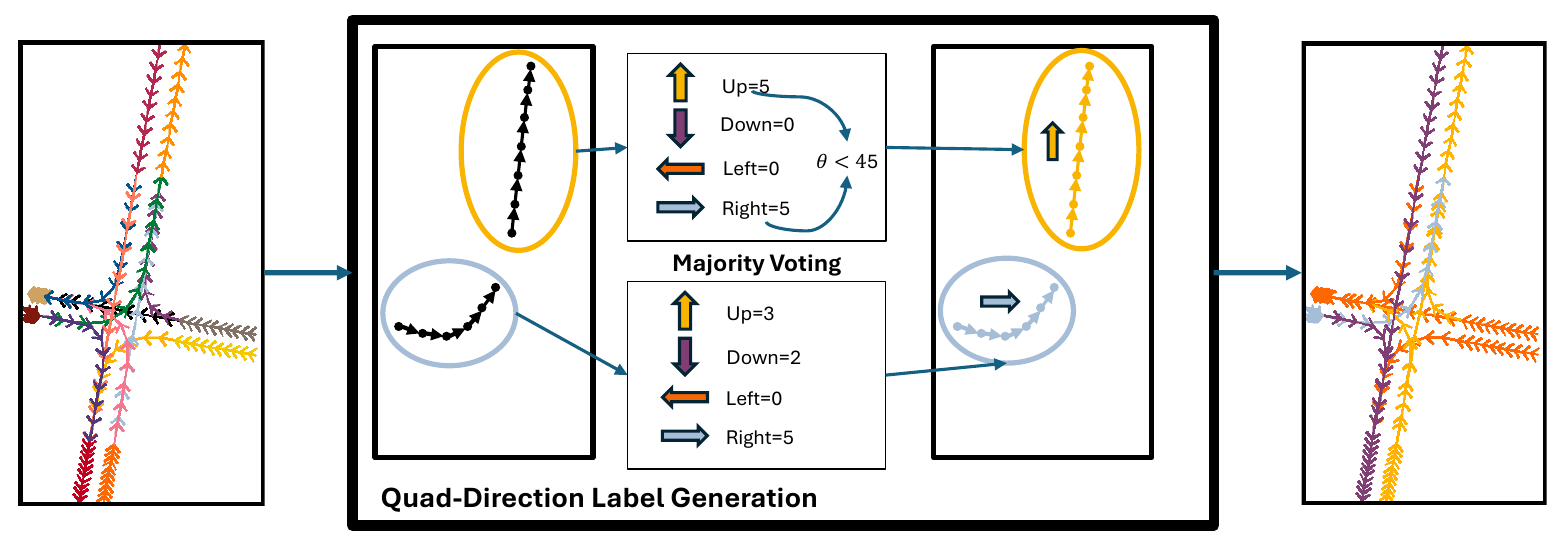}
  \vspace{-0.3cm}
  \caption{\textbf{Quad-Direction Labels Encoding} This figure demonstrates the creation of quad-direction labels, which encode the flow information of centerlines. These labels are up, down, left, and right. To generate these labels, a voting mechanism is applied between consecutive centerline points. In cases of equal votes, the final decision is based on the angle between the start and end points of the centerline.}
  \label{fig: quad_direction_label_representation}
  \vspace{-0.5cm}
\end{figure}

In the context of the road topology problem, comprehending the flow information of each centerline is crucial. However, traditional mask representations lack this flow information. To address this limitation, we propose the quad-direction label representation, which endows instance masks with flow-aware functionality. The generated quad-direction labels are treated as semantic information for the flow-aware mask instances and are trained using a cross-entropy classification loss. For simplicity, the quad-direction labels are defined as up, down, left, and right. Figure \ref{fig: quad_direction_label_representation} illustrates the process of determining the quad-direction labels for each centerline instance. The input to this process is the centerline instances, and the output is the quad-direction labels, which are utilized as semantic information for the centerlines.

The quad-direction label representation utilizes majority voting to assign direction labels to each centerline, as illustrated in Figure \ref{fig: quad_direction_label_representation}. Initially, the direction votes between consecutive points of the same centerline instance are determined along each axis. Each consecutive point pair casts up or down votes on one axis and left or right votes on the other axis. Subsequently, majority voting is conducted across all directions to ascertain the final quad-direction label. In instances where the number of votes from the two axes is equal, the direction label is assigned based on the angle between the start and end points of the centerline.

\vspace{-0.3cm}
\subsection{Quad-direction Label Aware Post Processing Technique}
\vspace{-0.1cm}
\label{sec: quad_direction_label_aware_post_processing_stage}

\begin{figure}[tb]
  \centering
  \includegraphics[width=0.7\linewidth]{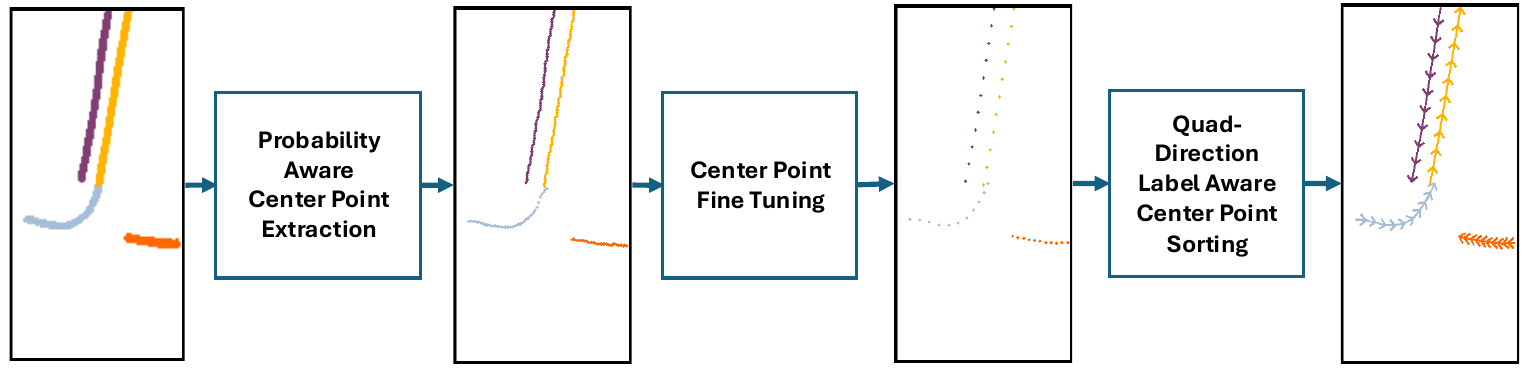}
  \vspace{-0.3cm}
  \caption{\textbf{Quad-Direction Label-Aware Post-Processing} The figure provides an overview of the technique, which converts flow-aware instance masks into centerlines. During the first and second stages of this technique, the mask representation is transformed into sparse BEV points. In the third stage, these points are ordered using quad-direction labels, thereby encoding the flow information of the centerlines.}
  \label{fig: quad_label_aware_post_process}
  \vspace{-0.5cm}
\end{figure} 

The TopoMask architecture aims to propose a flow-aware mask representation for each centerline. Each centerline is defined as a sequence of ordered 3D points, where the 3D points denote precise locations, and the order of the points indicates flow information. Although the flow-aware mask representation encodes this information, it does not precisely represent the centerline. A post-processing stage is then used to transform the flow-aware mask representation into a centerline representation. This proposed quad-direction label aware post-processing part consists of three stages, as shown in Figure \ref{fig: quad_label_aware_post_process}. 

In the first stage, called probability-aware center point extraction, row-wise and column-wise BEV location expectations are calculated from the mask probability maps of the BEV domain, which is inspired from \cite{qin2022ultra}. Assuming that the BEV mask prediction size is H$\times$W, there is a probability map of mask prediction $\mathbfcal{P}(x, y)$ for every centerline instance where $x$ and $y$ denote the vertical and horizontal positions, respectively. From this probability map, depending on the masking threshold $p$, a foreground mask $\mathbfcal{M}(x, y)$ is calculated. In the row-wise BEV location expectation implementation, the corresponding $y$ coordinate $\mathcal{R}_x$ for every $x$ coordinate (row) is calculated by
\begin{align*}
     \mathcal{R}_x = \sum_{y=0}^{W}  \frac{\mathbfcal{P}(x,y) \mathbfcal{M}(x,y) y } {\mathbfcal{P}(x,y) \mathbfcal{M}(x,y)}, 
\end{align*}
and the resulting point $\mathbb{P}_{\mathcal{R}_x} = (x, \mathcal{R}_x)$. The score of the extracted point $\mathcal{S}_{\mathcal{R}_x}$ is $\mathbfcal{P} \left ( x, \lfloor \mathcal{R}_x\rfloor \right )$
, where $\mathbfcal{P}(x, y)$ is the probability map and $\lfloor a \rfloor$ is the floor operation and preferred for simplicity. However, to accept the $\mathbb{P}_{\mathcal{R}_x}$ as a valid point, the condition of $\left( \mathcal{S}_{\mathcal{R}_x} > p \right )$ should be satisfied. Similarly, the column-wise BEV location expectation can estimate the point $\mathbb{P}_{\mathcal{C}_y}$, the score $\mathcal{S}_{\mathcal{C}_y}$ and the validity $\mathcal{V}_{\mathcal{C}_y}$. Benefitting from the quad-direction labels (see Section \ref{sec: quad_direction_label_representation}), row-wise BEV location expectation is applied for up/down directions, and column-wise BEV location estimation for left/right directions instead of utilizing both for all directions, leading to better computational efficiency.  

The second stage involves center point fine-tuning. In this stage, polynomial fitting and arc interpolation techniques are employed to enhance accuracy and reduce the number of points. Polynomial fitting mitigates noise in outliers and localization errors in segmentation to some extent. Arc interpolation ensures equally spaced points in the BEV domain. Sparsification is then applied to remove redundant information in the point set, which is essential for efficiency.

The third stage is quad-direction label-aware center point sorting. The output of the second stage consists of unordered point sets. In this stage, point sets are ordered according to the predicted quad-direction labels, which encode the flow information. For up and down direction labels, the points in a point set are ordered along one axis (the vehicle’s $x$ axis in its coordinate system). Conversely, for left and right direction labels, the points are ordered along the other axis (the vehicle’s $y$ axis in its coordinate system).

\vspace{-0.3cm}
\subsection{Multi-height Bin Implementation in Lift Splat Technique}
\vspace{-0.2cm}
\label{sec: multi_height_bin_implementation_in_lift_splat_technique}

\begin{figure}[tb]
  \centering
  \includegraphics[width=0.7\linewidth]{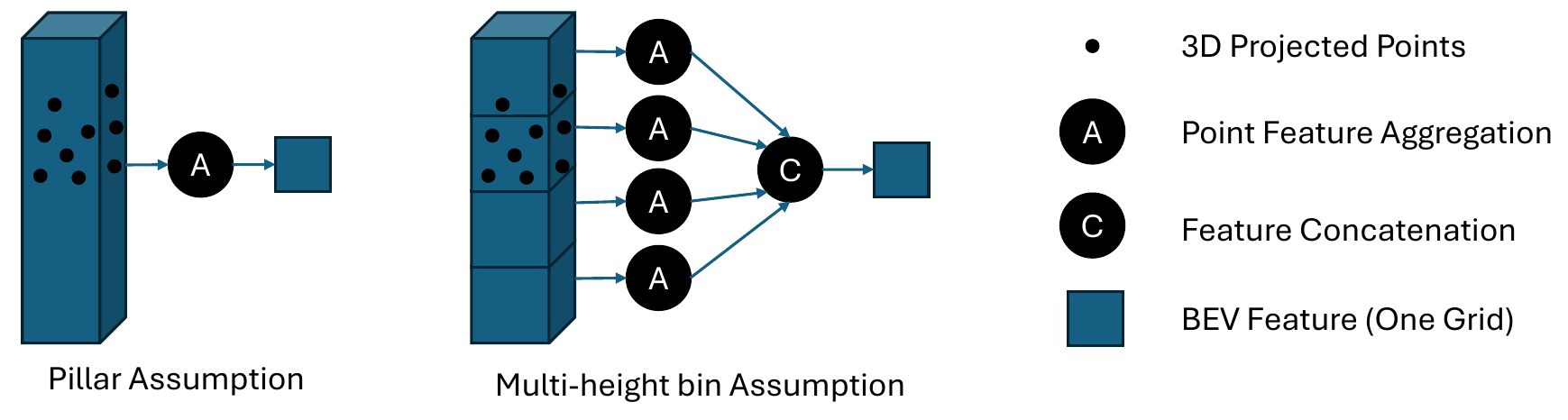}
  \vspace{-0.3cm}
  \caption{\textbf{Voxel Feature Aggregation Comparison}. The figure demonstrates the voxel aggregation techniques under the regular pillar assumption and the multi-height bin assumption. The proposed multi-height bin assumption encodes more information by preserving the practicality of the regular pillar assumption.}
  \label{fig: pillar_vs_multi_height_bin}
  \vspace{-0.5cm}
\end{figure}

The Lift Splat (LSS) technique is employed for BEV feature extraction in the TopoMask architecture (see Figure \ref{fig: topomask_architecture_centerline}). In this stage, LSS \cite{Philion2020} converts multi-camera image features into BEV features. The LSS technique projects pixels into the 3D world coordinate system through depth estimation, and the voxel pooling algorithm aggregates projected 3D world coordinates, converting them into BEV features. Studies utilizing LSS, such as BevDet \cite{Huang2021} and BevDepth \cite{Wang}, adhere to the vertical columns (pillars) assumption and do not employ multiple height bins (see Figure \ref{fig: pillar_vs_multi_height_bin}). However, it can be argued that the pillar assumption is inadequate for efficiently encoding voxel space, as height information is lost during the pooling stage.

Our study proposes enhancing centerline prediction performance by transforming the voxel pooling structure in the LSS technique into multiple bins. The outputs from these bins are then concatenated in the channel dimension, allowing for improved polyline prediction while retaining the practicality of the pillar assumption. The difference between the pillar assumption and multi-height bin implementation is illustrated in Figure \ref{fig: pillar_vs_multi_height_bin}. Furthermore, this study examines the optimality of the current lower and upper height limits of voxel space in the literature and explores better height limits for centerline prediction.

\vspace{-0.3cm}
\subsection{Fusion of Mask and Bezier Outputs} 
\vspace{-0.2cm}
\label{sec: fusion_of_mask_and_bezier_heads}

Fusion block utilizes the output of Bezier head and quad-direction label aware post-processing block (See Figure \ref{fig: topomask_architecture_centerline}). The proposed fusion block is simple, but it is important to understand that the Bezier head and the proposed mask head have complementary information to each other. After the post process of Bezier head and mask head outputs, both have N number of ordered point sets predictions, such that $\mathbf{b}_i=(x_i, y_i, z_i)$ is the $i$-th point prediction of Bezier head and $\mathbf{m}_i=(x_i, y_i, 0)$ is the $i$-th point prediction of mask head, denoting that mask head do not have height prediction mechanism. The fusion output is simply the average of both heads, which are 
\begin{align}
    \mathbf{f}_i=(\frac{\mathbf{m}_{x_i} + \mathbf{b}_{x_i}}{2}, \frac{\mathbf{m}_{y_i} + \mathbf{b}_{y_i}}{2}, \mathbf{b}_{z_i}).
\end{align}
\vspace{-0.7cm}

\vspace{-0.2cm}
\section{Experimental Evaluation}
\vspace{-0.2cm}
\subsection{Dataset and Metrics} 
\vspace{-0.2cm}

The OpenLane-V2 dataset \cite{wang2023openlanev2} comprises two subsets: Subset-A and Subset-B. Subset-A is derived from the Argoverse 2 (AV2) \cite{argoverse2} dataset, while Subset-B is based on the NuScenes \cite{nuscenes} dataset. A significant limitation of Subset-B is the absence of ground height information. For the evaluation of centerlines and traffic elements, the considered area spans +50 to -50 meters forward and +25 to -25 meters sideways, corresponding to the x and y directions in the vehicle coordinate system, respectively.

\begin{equation}
\label{Eq:OLS}
    \text{OLS} = \frac{1}{4} \bigg[ \text{DET}_{l} + \text{DET}_{t} + f(\text{TOP}_{ll}) + f(\text{TOP}_{lt}) \bigg]
\end{equation}

The evaluation system utilizes the OpenLane-V2 Score (OLS) as the overall metric, calculated as the average of multiple task metrics (Eq. \ref{Eq:OLS}): centerline prediction with Frechet-based mAP ($\text{DET}_{l}$), traffic element prediction with IOU-based mAP ($\text{DET}_{t}$), and topology relations between centerlines and traffic elements ($\text{TOP}_{ll}$ and $\text{TOP}_{lt}$). It is important to note that the evaluation metric pipeline has been updated from V1.0 to V1.1 to address identified issues, as highlighted in the TopoMLP study \cite{wu2023topomlp}. This study also employs Chamfer-based mAP ($\text{DET}_{l_{ch}}$) for a more in-depth analysis. The true positive criteria thresholds for $\text{DET}_{l_{ch}}$ are 0.5, 1, and 1.5 meters, while for $\text{DET}_{l}$, they are 1, 2, and 3 meters.

\vspace{-0.3cm}
\subsection{Experimental Details}
\vspace{-0.2cm}
The proposed architecture employs distinct backbones for the traffic elements and centerline branches, without weight sharing between them. This separation allows the traffic element branch to utilize different input resolutions and augmentation strategies. A multi-scale augmentation technique \cite{Zhu2020, Liu2022_dab} is adopted for training the traffic element branch. A deformable attention-based detection transformer, DAB-DETR \cite{Liu2022_dab}, is used to extract 2D traffic element queries. The denoising training strategy from DN-DETR \cite{li2022dndetr} and a two-stage structure inspired by DINO \cite{Zhang2022_dino} are implemented in the 2D traffic element branch. Consistent with the literature, a 0.5$\times$ image resolution and ResNet50 architecture are used for both the traffic element and centerline branches, while ResNet18 is employed as the BEV backbone following the Lift-Splat technique, unless specified otherwise in the related experiments.

In the experiments, we used a batch size of 8 and a learning rate of $3\times10^{-4}$, with a 0.1 multiplication factor applied to both PV and BEV backbones. The AdamW optimizer, with a weight decay of $1\times10^{-2}$, was employed. A polynomial learning rate decay was adopted, featuring a decay factor of 0.9 and a warm-up period of 1000 iterations. To expedite training, an efficient CUDA implementation of the Voxel Pooling (VP) operation \cite{huang2022bevpoolv2} was utilized in the Lift-Splat technique. Gradient norm clipping was set to 35.

Our masked-attention-based transformer decoder was designed over the study of Mask2former \cite{cheng2022masked}. The overlap threshold in the post-processing part of the Mask2Former is disabled (set to zero) to enhance performance. The threshold for point probability selection, following mask-to-point conversion, is established at 0.95 to optimize performance for the mask head. The BEV feature input to Mask2Former uses a grid size of $200\times104$ with a 0.5-meter grid interval. The width of the centerline instance is set to 4 during the conversion of the point set into the mask structure. Feature scale selection in the Mask2Former is adjusted to 1x, 0.5x, and 0.25x scale features, instead of lower scales, because BEV image resolutions are lower than PV image resolutions, and the computational increase is tolerable due to the higher performance. The number of control points is set to 4, and the number of queries is set to 200, as these configurations showed the best performance in the ablations. To further enhance the performance of the Mask2Former, we adopt a hybrid matching scheme \cite{jia2023detrs}, which integrates the original one-to-one matching with an auxiliary one-to-many matching concept.

\vspace{-0.3cm}
\subsection{Output Type Ablations} 
\vspace{-0.2cm}

\begin{table}[!tb]
    \caption{The performance of different output types in V1.1 topology evaluation. The table is categorized into two categories: without height prediction and with height prediction. Evaluations labeled as without height aim to measure pure BEV performance. }
    \label{tab: fusion_analysis}
    \vspace{-0.2cm}
    \centering
    \scalebox{0.8}{
    	\begin{tabular}{| c | c | c | c | c | c | c | c | c |}
                \toprule
                Output & \multicolumn{4}{c|}{Without Height}  & \multicolumn{4}{c|}{With Height} \\
                Type & DET$_{l}$  & DET$_{l_{ch}}$
                & TOP$_{ll}$ & TOP$_{lt}$ & DET$_{l}$  & DET$_{l_{ch}}$
                & TOP$_{ll}$ & TOP$_{lt}$ \\
                \midrule
                Mask & 30.3 & 29.8 & 20.3 & 31.3 & 30.8 & 36.1 & 21.2 & 32.6 \\
                Bezier & 30.4 & 30.1 & 21.1 & 31.6 & 31.8 & 36.8 & 22.1 & 33.0 \\
                Fusion & \textbf{31.7} & \textbf{30.8} & \textbf{22.6} & \textbf{33.1} & \textbf{34.4} & \textbf{37.7} & \textbf{23.8} & \textbf{34.4}\\
                \bottomrule
    	\end{tabular}
    }
    \vspace{-0.5cm}
\end{table} 

Table \ref{tab: fusion_analysis} presents the performance of various output types of the TopoMask architecture: mask, Bezier, and fusion. The fusion implementation is detailed in Section \ref{sec: fusion_of_mask_and_bezier_heads}. Notably, the mask head of the TopoMask architecture lacks a height prediction structure; hence, the table also includes performance results for Bezier and fusion outputs without height prediction. Additionally, the \textit{mask with height} section illustrates the performance of the mask output when enriched with height information from the Bezier output. The proposed mask output structure achieves performance levels very close to those of the Bezier output in both scenarios— with and without height prediction. This finding suggests that the mask output structure is a viable alternative to the Bezier output structure. Furthermore, the fusion of Bezier and mask outputs enhances performance across all metrics for both output types. This improvement indicates that the mask and Bezier outputs provide complementary information, and even a basic fusion algorithm can enhance both baselines.

\vspace{-0.3cm}
\subsection{Attention Type Ablations}
\vspace{-0.2cm}

\begin{table}[!tb]
    \caption{Performance results of TopoMask architecture concerning the attention types. }
    \label{tab: attention_type_ablations}
    \vspace{-0.3cm}
    \centering
    \scalebox{0.8}{
        \begin{tabular}{| c | ccccc |}
        \hline
        Attention Type  & DET$_l$ & DET$_{l_{ch}}$ & F-Score & PQ & TOP$_{ll}$  \\
        \midrule
        Standard Cross Attention  & 29.9 & 32.9 & 53.7 & 27.9 & 21.3  \\
        Deformable Attention  & 30.7 & 36.6 & 54.3 & 28.9 & 23.4 \\
        Masked Attention  & \textbf{32.2} & \textbf{37.2} & \textbf{56.3} & \textbf{30.6} & \textbf{24.1} \\
        \bottomrule
        \end{tabular}
    }
    \vspace{-0.2cm}
\end{table} 

The performance of the TopoMask architecture with various attention mechanisms is detailed in Table \ref{tab: attention_type_ablations}. In these experiments, only the attention mechanism is altered, while other components of the architecture remain constant, and the hybrid matching scheme \cite{jia2023detrs} is disabled due to the significant increase in computational complexity for standard cross attention. For all of the experiments, an instance-mask-based Hungarian matching strategy has been followed. To address convergence issues with deformable attention, the mask prediction head is modified to an iterative addition formulation akin to the Bezier keypoint regression (See Figure \ref{fig: topomask_masked_attention_detail}). For the reference points in the deformable attention mechanism, lane masks are converted to target boxes, and an additional box loss is trained alongside other losses. For further evaluation, the OpenLane-V2 F-score \cite{wang2023openlanev2}, and BEV Panoptic Quality \cite{Hu2021} metrics are included in the table. The results indicate that masked attention consistently outperforms across all metrics.

\vspace{-0.3cm}
\subsection{Multiple-height Bin Analysis}
\vspace{-0.2cm}

\begin{table}[!tb]
    \caption{Performance results concerning different height bin configurations in V1.1 metric baseline. Height bin configuration is denoted by (lower bound, upper bound, bin length).  }
    \label{tab: multi_height_bin_analysis}
    \vspace{-0.5cm}
    \centering
    \scalebox{0.8}{
        \begin{tabular}{| c | c | ccccc |}
        \hline
        Height Bin Configuration & \# Bins & DET$_l$ & DET$_{l_{ch}}$ & TOP$_{ll}$  & TOP$_{lt}$ & OLS \\
        \midrule
        (-5, 3, 8) & 1 & 31.2 & 34.2 & 20.8 & 31.8 & 45.9 \\
        (-5, 3, 2) & 4 & 31.5 & 35.2 & 21.6 & 32.5 & 46.9 \\
        (-5, 3, 1) & 8 & 32.1 & 37.3 & 22.6 & 32.9 & 46.7 \\
        \midrule
        (-5, 5, 1) & 10 & 32.7 & \textbf{37.9} & 23.4 & 34.0 & 47.9 \\
        (-10, 10, 1) & 20 & \textbf{34.4} & 37.3 & \textbf{23.7} & \textbf{34.7} & \textbf{48.6} \\
        \bottomrule
        \end{tabular}
    }
    \vspace{-0.5cm}
\end{table}

Multi-height bin implementation replaces the pillar implementation of voxel pooling in the LSS algorithm, details of which are given in Section \ref{sec: multi_height_bin_implementation_in_lift_splat_technique}. Performance analysis of multi-height bin implementation are provided in Table \ref{tab: multi_height_bin_analysis}. The default height bin configuration of LSS \cite{Philion2020}, BevDet \cite{Huang2021}, BevDepth \cite{Wang} is (-5, 3, 8), in which -5 denotes the lower height bound, 3 denotes the upper bound, and 8 denotes the bin length. This is a pillar implementation (single bin) as the difference between the lower and upper bound is also 8. As seen in the table, decreasing the bin length by preserving the lower and upper bound, which corresponds to an increase in the number of bins, increases the performance across all performance indicators. Additionally, maintaining the same bin length while increasing the upper bound from 3 to 5 improves performance across all indicators. Further changing both the lower and upper bounds to -10 and 10, respectively, while keeping the bin length constant, enhances performance for all metrics except the DET$_{l_{ch}}$. However, since DET$_l$ in OLS is considered the primary performance metric, (-10, 10, 1) configuration is set as the default.

\vspace{-0.3cm}
\subsection{Comparative Evaluations}
\vspace{-0.2cm}\label{sec: comparative_evaluations}

\begin{table}[!tb]
    \caption{Comparative evaluation of TopoMask with the other methods on Subset-A with regards to both V1.0 and V1.1 metrics. }
    \label{tab: sota_A}
    \vspace{-0.3cm}
    \centering
    \scalebox{0.8}{
    	\begin{tabular}{| c | c c |cc|ccc|ccc|}
    		\toprule
                 &  &  &  &  & \multicolumn{3}{|c|}{V1.0} & \multicolumn{3}{|c|}{V1.1} \\
                 \midrule
                Method & Backbone & Epoch & DET$_{l}$ & DET$_{t}$
                & TOP$_{ll}$  & TOP$_{lt}$
                & OLS
                & TOP$_{ll}$  & TOP$_{lt}$
                & OLS
                \\
                \midrule
                STSU & ResNet50 & 24 & 12.7 & 43.0 & 0.5 & 15.1 & 25.4 & 2.9 & 19.8 & 29.3\\
                 VectorMapNet & ResNet50 & 24 & 11.1 & 41.7 & 0.4 & 6.2 & 20.8 & 2.7 & 9.2 & 24.9\\
                 MapTR & ResNet50 & 24 & 8.3 & 43.5 & 0.2 & 5.8 & 20.0 & 2.3 & 8.3 & 24.2\\
                 TopoNet & ResNet50 & 24 & 28.6 & 48.6 & 4.1 & 20.3 & 35.6 & 10.9 & 23.9 & 39.8\\
                 TopoMLP & ResNet50 & 24 & 28.5 & 49.5 & 7.2 & \textbf{23.4} & 38.3 & 21.7 & 26.9 & 44.1 \\
                 TopoMask & ResNet50 & 24 & \textbf{34.5} & \textbf{53.8} & \textbf{10.8} & 20.7 & \textbf{41.7} & \textbf{24.5} & \textbf{35.6} & \textbf{49.4} \\ 
                \bottomrule
    	\end{tabular}
    }
    \vspace{-0.0cm}
\end{table} 

\begin{table}[!tb]
    \caption{Comparative evaluation of TopoMask with other methods on the Subset-B. }
    \label{tab: sota_B}
    \vspace{-0.3cm}
    \centering 
    \scalebox{0.8}{
    	\begin{tabular}{| c | c c | cc|ccc|ccc|}
    		\toprule
                 &  &  &  &  & \multicolumn{3}{|c|}{V1.0} & \multicolumn{3}{|c|}{V1.1} \\
                 \midrule
                Method & Backbone & Epoch & DET$_{l}$ & DET$_{t}$
                & TOP$_{ll}$  & TOP$_{lt}$
                & OLS
                & TOP$_{ll}$  & TOP$_{lt}$
                & OLS
                \\
                \midrule
                 TopoNet & ResNet50 & 24 & 24.3 & 55.0 & 2.5 & 14.2 & 33.2 & 6.7 & 16.7 & 36.0 \\
                 TopoMLP & ResNet50 & 24 & 25.2 & \textbf{63.1} & 6.8 & \textbf{17.9} & 39.2 & 20.7 & 20.3 & 44.7 \\
                 TopoMask & ResNet50 & 24 &\textbf{41.6} & 61.1 & \textbf{12.4} & 14.2 & \textbf{43.9} & \textbf{28.7} & \textbf{26.1} & \textbf{51.8} \\
                \bottomrule
    	\end{tabular}
    }
    \vspace{-0.4cm}
\end{table}

Tables \ref{tab: sota_A} and \ref{tab: sota_B} present comparative results with the literature for Subset-A and Subset-B, respectively. The results for the TopoMask architecture are with fusion output type. For TopoMLP, we used the weights provided in the original repo to run the experiments, as V1.1 score performances are not reported in the paper. Results are reported using both versions of the OpenLane-V2 evaluation method. V1.1 addresses a bug found in V1.0 and provides a more accurate representation of topology performances. According to these tables, TopoMask outperforms all methods in all performance metrics in both V1.0 and V1.1, except TOP$_{lt}$ in V1.0. However, it can be argued that this is counterbalanced by the fact that it has a better TOP$_{lt}$ in V1.1, considering V1.1 is accepted to be a better measure of the topology performance. In Subset-B, TopoMLP obtained better DET$_{t}$ performance than TopoMask. However, this might not be a fair comparison as the resolution used in the original repo of TopoMLP is not clear and might be higher. TopoMLP study \cite{wu2023topomlp} indicates DET$_{t}$ as 53.3. In Subset-B, TopoMask significantly outperforms TopoMLP and TopoNet in DET$_{l}$ metric (41.6 vs 25.2).

\vspace{-0.3cm}
\section{Further Discussions}
\vspace{-0.3cm}
\label{sec: topology_related_modifications}

We argue that despite the upgrade from V1.0 to V1.1, current topology evaluation metrics still require further updates for fair comparison. The original OpenLane-V2 evaluation methodology uses a threshold of 0.5 for the relation confidence scores, which eliminates all predictions on the right half of the precision recall plot. On the other hand, a better threshold value may potentially yield better predictions by retaining those that would otherwise be eliminated with a fixed threshold value. 

\begin{table}[!tb]
    \caption{The analysis of topology confidence score manipulation on V1.1 evaluation system. The blue text indicates the increase with the topology confidence score manipulation technique.}
    \label{tab: score_boosting_analysis}
    \vspace{-0.3cm}
    \centering 
    \scalebox{0.8}{
    	\begin{tabular}{|c|ccc|ccc|}
    		\toprule
                 & \multicolumn{3}{|c|}{V1.1} & \multicolumn{3}{|c|}{V1.1 with Score Manipulation} \\
                 \midrule
                Method
                & TOP$_{ll}$ & TOP$_{lt}$
                & OLS$\uparrow$ 
                & TOP$_{ll}$$\uparrow$  & TOP$_{lt}$
                & OLS
                \\
                \midrule
                TopoNet & 10.9 & 23.8 & 39.8 & 21.5 \textcolor{blue}{(\textbf{+10.6)}} & 25.3 \textcolor{blue}{(\textbf{+1.5)}} & 43.5 \textcolor{blue}{(\textbf{+3.7)}} \\
                TopoMLP & 21.7 & 26.9 & 44.1 & 24.3 \textcolor{blue}{(+2.7)} & 27.1 \textcolor{blue}{(+0.2)} & 44.8 \textcolor{blue}{(+0.7)} \\
                TopoMask & \textbf{24.5} & \textbf{35.6} & \textbf{49.4} & \textbf{25.7} \textcolor{blue}{(+1.2)}& \textbf{36.7} \textcolor{blue}{(+1.1)} & \textbf{49.9} \textcolor{blue}{(+0.5)} \\
                \bottomrule
    	\end{tabular}
    }
    \vspace{-0.3cm}
\end{table} 

While mAP-like metrics coherently sort the scores of true positives and false positives, they require the entire precision-recall plot. To address this problem, we apply a confidence score manipulation strategy without any modifications to the evaluation baseline. Specifically, we increase the relation confidence scores by 1.0 for values exceeding the threshold of 0.05, such that $P(x) + 1 \times [P(x) > 0.05]$. This preserves the order of the scores and ensures that most predictions are included in the metric calculation. Based on this observation, we propose removing the confidence score thresholding mechanism from the original evaluation system to facilitate a fair comparison with the updated metric. 

The results of metric implementation with the topology confidence score manipulation are presented in Table \ref{tab: score_boosting_analysis}. For the TopoNet and TopoMLP results, the weights from the original repos are utilized. As indicated in the table, \textit{without changing anything in the evaluation baseline}, and \textit{by only remapping the scores of the topologies}, improvements are up to 10.6 in TOP$_{ll}$, 1.5 in TOP$_{lt}$ and 3.7 in OLS score.

\vspace{-0.3cm}
\section{Conclusions and Limitations}
\vspace{-0.3cm}
This study proposes the explicit use of instance mask formulation for centerline prediction as an alternative to parametric or keypoint-based methods. Additionally, we explore the fusion of outputs from both parametric and instance-mask approaches to achieve state-of-the-art results. We also demonstrate that enhancing LSS with a multi-height bin implementation improves centerline detection performance. Our findings indicate the superiority of masked attention over standard deformable attention for centerline detection. Furthermore, we critically analyze current topology evaluation metrics and suggest potential improvements.

One limitation of mask-based approaches is their susceptibility to errors arising from the grid structure in the BEV domain. The thresholds in current centerline detection metrics are inadequate for probing these errors, necessitating additional metrics for detailed localization performance. Additionally, the proposed method lacks a proper height prediction module for the mask head. Therefore, enhancing TopoMask heads with additional heads could lead to further improvements.

\vspace{-0.3cm}
\section*{\small ACKNOWLEDGMENTS}
\vspace{-0.4cm}
{\small The numerical calculations reported in this paper were partially performed at TUBITAK ULAKBIM, High Performance and Grid Computing Center (TRUBA resources)}

\clearpage  

%
%
\bibliographystyle{splncs04}
\bibliography{main}

\end{document}